\documentclass[11pt,a4paper]{article}
\usepackage{acl2014}
\usepackage{times}
\usepackage{latexsym}
\usepackage{url}
\usepackage{color}
\usepackage{multirow}
\usepackage{comment}
\usepackage{amsmath} 
\usepackage{amssymb}

\usepackage{etoolbox}
\usepackage{graphicx}
 
 \usepackage{enumitem}
  {\begin{itemize}[topsep=0pt, partopsep=0pt] \footnotesize%
    \setlength{\itemsep}{0pt}%
    \setlength{\parskip}{0pt}%
    }%
  {\end{itemize}}
  
  \usepackage{color}
  {\begin{enumerate}≈ \footnotesize%
    \setlength{\itemsep}{0pt}%
    \setlength{\parskip}{0pt}%
    }%
  {\end{enumerate}}
  
  \usepackage{CJKutf8}
  
  \newcommand{\argmax}{\operatornamewithlimits{argmax}}
\usepackage{soul}
\title{Do Multi-Sense Embeddings Improve Natural Language Understanding?}
\author{Jiwei Li \\
{ Computer Science Department} \\
{ Stanford University}\\
{ Stanford, CA 94305, USA}\\
{ jiweil@stanford.edu}\And
Dan Jurafsky\\
{ Computer Science Department} \\
{ Stanford University}\\
{ Stanford, CA 94305, USA}\\
jurafsky@stanford.edu
}
\begin{document}
\maketitle
\begin{abstract}
%Recent vector models of word meaning 
Learning a distinct representation for each sense of an ambiguous word 
could lead to more powerful and fine-grained models of
vector-space representations. Yet while `multi-sense' methods have been proposed
and tested on artificial word-similarity tasks,
we don't know if they improve real natural language understanding tasks.
In this paper we introduce 
a multi-sense embedding model based on Chinese Restaurant Processes
that achieves state of the art performance on matching 
human word similarity judgments, and propose
a pipelined architecture for
incorporating multi-sense embeddings into language understanding.

We then test the performance of our
model on part-of-speech tagging, named
entity recognition, sentiment analysis, semantic relation identification and semantic relatedness,
controlling for embedding dimensionality.
We find that multi-sense embeddings do improve performance on
some tasks (part-of-speech tagging, semantic relation identification, semantic relatedness)
but not on others (named entity recognition, various forms of sentiment analysis).
We discuss how these differences may be caused by the different role
of word sense information in each of the tasks.  The results highlight the importance of testing embedding models
in real applications.

%introduce slight performance boost in semantic-related tasks, but is of little in others that
%depend on correctly identifying a few key words
 %such as sentiment analysis.

\end{abstract}
\section{Introduction}
Enriching vector models of word meaning so they can represent multiple word
senses per word type seems to offer the potential to improve many language understanding tasks. 
Most traditional embedding models associate each word type  with  a single embedding (e.g., \newcite{bengio2006neural}). Thus the embedding for homonymous words like {\em bank}  
(with senses including `sloping land' and `financial institution') 
is forced to represent some uneasy central tendency between the various meanings.
More fine-grained embeddings that represent more natural regions in semantic space
could thus improve language understanding.

Early research pointed out that embeddings could model aspects of word sense
\cite{kintsch2001predication}
and recent research has proposed a number of models that
represent each word type by different senses,
each sense associated with a sense-specific embedding
\cite{kintsch2001predication,reisinger2010multi,neelakantan2014efficient,huang2012improving,chen2014unified,pina2014simple,wu2015sense,liu2015topical}.  
Such sense-specific embeddings have shown 
improved performance on simple artificial tasks like matching human
word similarity judgments--- WS353 \cite{rubenstein1965contextual} or MC30 \cite{huang2012improving}. 

Incorporating multisense
word embeddings into general NLP
tasks requires a pipelined architecture that addresses three major steps:
\begin{enumerate}
\item {\bf Sense-specific representation learning}:
learn word sense specific embeddings
from a large corpus, either unsupervised or
aided by external resources like WordNet.
\item {\bf Sense induction}: given a text unit (a
phrase, sentence, document, etc.), infer
word senses for its tokens and associate them with corresponding sense-specific embeddings.
\item {\bf Representation acquisition for
phrases or sentences:} learn representations
for text units given sense-specific
embeddings and pass them to
machine learning classifiers.
\end{enumerate}

Most existing work on multi-sense embeddings emphasizes
the first step by learning sense specific embeddings,
but does not explore the next two steps. These are important steps, however,
since it isn't clear how existing
multi-sense embeddings can be incorporated into and benefit
real-world NLU tasks.

We propose a pipelined architecture to address all three
steps and apply it to a variety of
NLP tasks: part-of-speech tagging, named entity
recognition, sentiment analysis, semantic relation
identification and semantic relatedness.  We find:

\begin{itemize}
\item Multi-sense embeddings give
improved performance in some tasks
(e.g., semantic similarity for words and sentences, semantic relation identification
part-of-speech tagging),
but not others (e.g., sentiment analysis, named entity extraction).
In our analysis we offer some suggested explanations for these differences.
\begin{comment}
and that concatenating global (one-sense-per-vector)
embeddings with sense-specific embeddings
gives significantly improved performance.
But none of these differences are significant
when carefully compared with embeddings
of the same dimensionality
\end{comment}
\item Some of the improvements for multi-sense embeddings are no longer visible
when using more sophisticated neural models
like LSTMs which have more flexibility in filtering
away the informational chaff from the wheat.
%\item Multi-sense embeddings offer little help in
%tasks that depend on correctly identifying a
%few key words like sentiment analysis.
\item It is important to carefully compare
against embeddings of the same dimensionality.
\item When doing so, the most straightforward
way to yield better performance on these
tasks is just to increase embedding dimensionality.
\end{itemize}

After describing related work, we introduce the new unsupervised sense-learning
model in section 3, give our sense-induction algorithm in section 4, and then
in following sections evaluate its performance for word similarity, and then various NLP tasks.

\section{Related Work}
Neural embedding learning frameworks represent each token with a dense vector representation,
optimized through predicting neighboring words or decomposing co-occurrence matrices
\cite{bengio2006neural,collobert2008unified,mnih2007three,mikolov2013efficient,mikolov2010recurrent,pennington2014glove}. Standard neural models represent each word with a single unique vector representation. 

Recent work has begun to augment the neural paradigm to 
address the multi-sense problem by associating each word with a series of sense specific
embeddings. The central idea is to augment standard
embedding learning models like skip-grams by disambiguating word
senses based on local co-occurrence--- e.g., the fruit ``apple" tends
to co-occur with the words ``cider, tree, pear" while the homophonous IT company
co-occurs with words like ``iphone", ``Google" or ``ipod".

For example \newcite{reisinger2010multi} and \newcite{huang2012improving}
propose ways to develop multiple embeddings per word type
by pre-clustering the contexts of each token to create a fixed number of senses for each word,
and then relabeling each word token with the clustered sense
before learning embeddings.
\newcite{neelakantan2014efficient} extend these models by relaxing the assumption
that each word must have a fixed number of senses and using a non-parametric model
setting a threshold to decide when a new sense cluster should be split off;
\newcite{liu2015topical}
learns sense/topic specific embeddings by combining neural frameworks with LDA topic models.
\newcite{wu2015sense} disambiguate sense embeddings from Wikipedia by first clustering wiki documents.  
\newcite{chen2014unified} turn to external resources and used a predefined inventory of senses, building a 
distinct representation for every sense defined by the Wordnet dictionary. 
Other relevant work  includes \newcite{qiu2014learning} who maintains separate representations for different part-of-speech tags of the same word. 

Recent work is mostly evaluated  on the relatively
artificial task of matching human word
similarity judgments.
%rather than on exploring whether multi-sense embeddings can improve natural
%language understanding in real-world tasks.

\section{Learning Sense-Specific Embeddings}

We propose to build on this previous literature,
most specifically \newcite{huang2012improving} and \newcite{neelakantan2014efficient},
to develop an algorithm for learning multiple embeddings for
each word type, each embedding corresponding to a distinct induced word sense.
Such an algorithm should have the property that a word should be associated with a new sense vector 
just when evidence in the context 
(e.g., neighboring words, document-level co-occurrence statistics) 
suggests that it is sufficiently different from its early senses.
Such a line of thinking naturally points to Chinese Restaurant Processes (CRP)
\cite{griffiths2004hierarchical,teh2006hierarchical} which have been applied in the related field of word sense induction.
In the analogy of CRP, the current word could either  sit at one of the existing tables
(belonging to one of the existing senses) or choose a new table (a new sense). 
The decision is made by measuring semantic relatedness (based on local context information and
global document information) and the number of customers already sitting at that table (the popularity of word senses). 
We propose such a model and show that it improves over the state of the art on a standard word similarity task.

\subsection{Chinese Restaurant Processes}
We offer a brief overview of Chinese Restaurant Processes in this section; readers
interested in more details can consult the original papers
\cite{griffiths2004hierarchical,teh2006hierarchical,pitman1995exchangeable}. 
CRP can be viewed as a practical interpretation of Dirichlet Processes \cite{ferguson1973bayesian} for non-parametric clustering. 
In the analogy, each data point is compared to a customer in a restaurant. 
The restaurant has a series of tables $t$, each of which serves a dish $d_t$. This dish can be viewed as the index of a cluster or a topic. 
The next customer $w$ to enter would either choose an existing table, sharing the dish (cluster) already served or choosing a new cluster based on the following probability distribution: 
\begin{equation}
\begin{aligned}
Pr(t_w=t)
\propto \left\{
\begin{aligned}
&N_t P(w|d_t)~~~~~\text{if t already exists}\\
&\gamma P(w|d_{new})~~~\text{if t is new}
\end{aligned}
\right.
\end{aligned}
\end{equation}
where $N_t$ denotes the number of customers already sitting at table $t$ and $P(w|d_{t})$ denotes the probability of assigning the current data point to cluster $d_t$. $\gamma$ is the hyper parameter controlling the preference for sitting at a new table. 

CRPs exhibit  a useful ``rich get richer'' property because they take into account the popularity of different word senses. 
They are also more flexible than a simple threshold strategy for setting up new clusters,
due to the robustness introduced by adopting the relative ratio of  $P(w|d_t)$ and $P(w|d_{new})$.

\subsection{Incorporating CRP into Distributed Language Models}
We describe how we incorporate CRP into a standard distributed language model\footnote{We omit details about training standard distributed models; see \newcite{collobert2008unified} and \newcite{mikolov2013efficient}.}. 

As in the standard vector-space model,
each token $w$ is associated with a K dimensional global embedding $e_w$.
Additionally, it is associated with  a set of senses $Z_w=\{z_w^1, z_w^2, ..., z_w^{|Z_w|}\}$
where $|Z_w|$ denotes the number of senses discovered for word $w$. 
Each sense $z$ is associated with a distinct sense-specific embedding $e_w^z$.
When we encounter a new token $w$ in the text, at the first stage, we 
maximize the probability of 
seeing
the current token 
given its context
as in standard language models using the global vector $e_w$:
\begin{equation}
\begin{aligned}
p(e_w| e_{\text{neigh}})=F (e_w, e_{\text{neigh}})
\end{aligned}
\label{4}
\end{equation}
F() can take different forms in different learning paradigms, e.g., $F=\prod_{w'\in neigh} p(e_w, e_{w'})$ for skip-gram or $F=p(e_w, g(e_w))$ for SENNA \cite{collobert2008unified} and CBOW, where $g(e_{neigh})$ denotes a function that projects the concatenation of neighboring vectors to a vector with the same dimension as $e_w$ for SENNA and the bag-or-word averaging for CBOW \cite{mikolov2013efficient}.

Unlike traditional one-word-one-vector frameworks,
$e_{\text{neigh}}$ 
includes sense information in addition to the global vectors for neighbors.
$e_{\text{neigh}}$ can therefore be written as\footnote{For models that predict succeeding words, sense labels for preceding words have already been decided. For models that predict words using both left and right contexts, the labels for right-context words  have not been decided yet. In such cases we just use its global word vector to fill up the position.}.
\begin{equation}
\begin{aligned}
&e_{\text{neigh}}=\{e_{n-k},,..., e_{n-1}, e_{n+1},..., e_{n-k}\}
\end{aligned}
\label{3}
\end{equation}
\begin{comment}
$[e_{n-1},...,e_{n-k}, e_{n+1},.., e_{n+k}]$.
\end{comment}

 Next we would use CRP to decide which sense the current occurrence corresponds to, or construct a new sense  if it is a new meaning that we have not encountered before.
Based on CRP, the probability that assigns the current occurrence to each of the discovered senses or a new sense is given by:
\begin{equation}
\begin{aligned}
Pr(z_w=z)
\propto \left\{
\begin{aligned}
&N_z^w P(e_w^z|\text{context})~~\\
&~~~~~~~~~~~\text{if z already exists}\\
&\gamma P(w|z_{new})~~~\text{if z is new}
\end{aligned}
\right.
\end{aligned}
\label{4}
\end{equation}
where $N_z^w$ denotes the number of times already assigned to sense $z$ for token $w$. 
$P(e_w^z|\text{context})$ denotes the probability that current occurrence belonging to (or generated by) sense z.

The algorithm for parameter update for the one token predicting procedure is illustrated in Figure \ref{algorithm}: 
 Line 2 shows parameter updating through predicting the occurrence of current token.
Lines 4-6 illustrate the situation when a new word sense is detected, in which case we would add the newly detected sense $z$ into $Z_{w_n}$. 
The vector representation $e_{w}^z$ for the newly detected sense would be obtained by maximizing the function $p(e_w^z|\text{context})$.  
 
\begin{figure}[t!]
\rule{8cm}{0.03cm}
01: {\bf Input} : Token sequence $\{w_n,w_{\text{neigh}}\}$. \\
02: Update parameters involved in Equ (\ref{3})(\ref{4}) based on current word prediction.\\
03:~~ Sample sense label $z$ from CRP.\\
04: ~~If a new sense label $z$ is sampled: \\
05: \hspace{0.6cm} - add $z$ to $Z_{w_n}$\\
06: \hspace{0.6cm} - $e_{w_n}^z=\argmax p(w_n|z_m)$\\
07:~~else: update parameters involved based on sampled sense label z. \\
\rule{8cm}{0.03cm}
\caption{Incorporating CRP into Neural Language Models.}\label{algorithm}
\end{figure}

As we can see,
the model performs word-sense clustering and embedding learning jointly,
each one affecting the other.
The prediction of the global vector of the current token (line2) is based on 
both the global and sense-specific embeddings of its neighbors, as will be
updated through predicting the current token. 
Similarly, once the sense label is decided (line7), the model will adjust the embeddings for 
neighboring words, both global word vectors and sense-specific vectors.

\begin{comment}
and vector concatenation (VC). 
For VC, we use 0.5 dropout rate \cite{hinton2012improving}.
We iterate three times over the corpus.
\end{comment}

\section{Obtaining Word Representations for NLU tasks}
Next we describe how we decide sense labels for tokens in context. 
The scenario is treated as a inference procedure for sense labels where all global word embeddings and
sense-specific embeddings are kept fixed.

Given a document or a sentence, we  have an objective function with respect to sense labels by multiplying Eq.2 over each containing token. 
Computing the 
global optimum sense labeling---in which every word gets an optimal sense label---
requires searching over the space of all senses for all words, which can be expensive.  
We therefore chose two simplified heuristic approaches: 
\begin{itemize}
\item {\bf Greedy Search}: Assign each token the locally optimum sense label 
and represent the current token with the embedding associated with that sense.
\item  {\bf Expectation}: Compute the probability of each possible sense for the current 
word, and represent the word with the expectation vector:
$$\vec{e}_w=\sum_{z\in Z_w}p(w|z,\text{context})\cdot e_w^z$$
\end{itemize}

\section{Word Similarity Evaluation}
We evaluate our embeddings by comparing with other multi-sense embeddings on 
the standard
artificial  task
 for matching human word similarity judgments.
 \begin{comment}
  The aim of doing so is to show how well embeddings trained from the proposed framework compared with other types of multi-sense embeddings, which  serves as a premise to extend them to NLU tasks\footnote{To note, as mentioned earlier, existing work only list sense-specific embeddings for different tokens. It is not unclear how the second step in the pipelined system 
(as described  in Section 4) could be executed with their embeddings.}. 
\end{comment}

Early work used
similarity datasets like WS353 \cite{finkelstein2001placing} or RG  \cite{rubenstein1965contextual},
whose context-free nature makes them a poor evaluation.
We therefore adopt Stanford's Contextual Word Similarities 
(SCWS) \cite{huang2012improving},
in which human judgments are associated
with pairs of words in context.   Thus for example
``bank" in the context of ``river bank" would have low relatedness with ``deficit" 
in the context ``financial deficit".   

%
%are not particularly fit for our scenario  with multi sense representation
%
%\footnote{\cite{neelakantan2014efficient} computes word similarity by averaging similarities between all word senses.}. 

We trained our models on the two datasets: Wikipedia dataset which is comprised of 1.1 billion tokens and 
a large dataset by combining Wikipedia, Gigaword and Common crawl dataset, which is comprised of 120 billion tokens. 
We iterate over the dataset for 3 times, with window size 11. 
We next use the Greedy or Expectation strategies to
obtain word vectors for tokens given their context.
These vectors are then used as input to get the value of cosine similarity between two words.

\begin{table}
\centering
\small
\begin{tabular}{ccc}\hline
Model&Dataset&SCWS Correlation\\\hline
SkipGram&1.1B (wiki)&64.6\\
SG+Greedy&1.1B (wiki)&66.4\\
SG+Expect&1.1B (wiki)&67.0\\\hline
SkipGram&120B &66.4\\
SG+Greedy& 120B&69.1\\
SG+Expect& 120B&69.7\\\hline\hline
\end{tabular}
\begin{comment}
VC&66.0&SG&64.8\\
VC+NoGlobal&69.2&SG+NoGlobal&69.0 \\
\end{comment}
\caption{Performances for different set of multi-sense embeddings (300d) evaluated on SCWS by measuring the Spearman correlation
between each model’s similarity and the human judgments.}
\label{SCWS}
\end{table}

Performances are reported in Table \ref{SCWS}. Consistent with earlier work (e.g.., \newcite{neelakantan2014efficient}), we find that
multi-sense embeddings result in better performance
in the context-dependent   SCWS task
(SG+Greedy and SG+Expect are better than SG).
As expected, performance is not as high 
when global level information is ignored when choosing
word senses (SG+Greedy)  as when it is included (SG+Expect),
as neighboring words don't provide sufficient information for word sense disambiguation.
SG+Expect  yields +2.4 performance boost than one-word-one-vector strategy on 1.1 billion Wikipedia dataset\footnote{ 
\cite{neelakantan2014efficient} reported a result of 69.3 on SCWS dataset
trained from Wikipedia corpus, outperforming the proposed model described in this paper trained on the similar-size corpus in spite of the fact that different Wikipedia dumps and preprocessing techniques are adopted.} and +3.2 on the common crawl datast. 

\begin{comment}
A word is then represented by the concatenation of its global word vector and sense-based vector (either computed by the Greedy or Expectation strategies).  Unless otherwise noted, we use embeddings obtained by using this strategy as the input to NLU tasks. 
\end{comment}

\paragraph{Visualization} Table \ref{tab-illustration} shows examples of semantically related words given the local context. Word embeddings
for tokens are obtained by using the inferred sense labels from the Greedy model
and are then used to search for nearest neighbors in the vector space based on cosine similarity. 
Like earlier models (e.g., \newcite{neelakantan2014efficient}).,
the model can disambiguate different word senses 
(in examples like {\em bank, rock} and {\em apple}) based on their local context;
although of course the model is also capable of dealing with polysemy---senses that are less distinct.

\begin{table*}[!htb]
\small
\centering
\begin{tabular}{|c|c|}
\hline
Context                                                                                              & Nearest Neighbors                                                                                              \\ \hline\hline
{\bf Apple} is a kind of fruit.                                                                            & pear, cherry, mango, juice, peach, plum, fruit, cider, apples, tomato, orange, bean, pie                       \\ \hline
{\bf Apple} releases its new ipads.                                                                        & microsoft, intel, dell, ipad, macintosh, ipod, iphone, google, computer, imac, hardware                        \\ \hline\hline
\multicolumn{1}{|l|}{He borrowed the money from {\bf banks}.}                                                 & \multicolumn{1}{l|}{banking, credit, investment, finance, citibank, currency, assets, loads, imf, hsbc}        \\ \hline
\begin{tabular}[c]{@{}c@{}}along the shores of lakes, \\ {\bf banks} of rivers\end{tabular}                & \multicolumn{1}{l|}{land, coast, river, waters, stream, inland, area, coasts, shoreline, shores, peninsula}    \\ \hline\hline
\multicolumn{1}{|l|}{Basalt is the commonest volcanic {\bf rock}.}                                         & \multicolumn{1}{l|}{boulder, stone, rocks, sand, mud, limestone, volcanic, sedimentary, pelt, lava, basalt}    \\ \hline
\multicolumn{1}{|l|}{{\bf Rock} is the music of teenage rebellion.}                                        & \multicolumn{1}{l|}{band, pop, bands, song, rap, album, jazz. blues, singer, hip-pop, songs, guitar, musician} \\ \hline\hline
\end{tabular}
\caption{Nearest neighbors of words given context. The embeddings from context words are first inferred with the Greedy strategy; nearest neighbors are computed by cosine similarity between word embeddings. Similar phenomena have been observed in earlier work \cite{neelakantan2014efficient}}
\label{tab-illustration}
\end{table*}

\begin{comment}
A {\bf queen} is a woman who rules a country                                                               & princess, king, elizabeth, royal, monarch, prince, majesty, throne, mother                                     \\ \hline
\begin{tabular}[c]{@{}c@{}}A {\bf queen} ant is a reproducing\\   female ant in an ant colony\end{tabular} & ant, bee, female, colony, offspring, wasp, bees, eusociality, insect, king, reproductive                              \\ \hline\hline
\end{comment}

\section{Experiments on NLP Tasks}

Having shown that multi-sense embeddings improve word similarity tasks, 
we turn to ask whether they improve 
real-world NLU tasks: POS tagging, NER tagging, sentiment analysis at the phrase and sentence level, semantic relationship identification and sentence-level semantic relatedness. 
For each task, we experimented on the following sets of embeddings, which 
are trained using the word2vec package on the same corpus: 
\begin{itemize}
\itemsep 0pt
\item Standard one-word-one-vector embeddings from skip-gram (50d). 
\item  Sense disambiguated embeddings from Section 3 and 4 using Greedy Search and Expectation (50d)
\item The concatenation of global word embeddings and sense-specific embeddings (100d).  
\item 
Standard one-word-one-vector skip-gram embeddings with dimensionality doubled (100d)
(100d is the correct corresponding baseline 
since  the concatenation above doubles the dimensionality of word vectors)
\item Embeddings with very high dimensionality (300d). 
\end{itemize}

As far as possible we try to perform  an apple-to-apple comparison on these tasks,
and our goal is an analytic one---to investigate how well semantic information can be encoded in 
multi-sense embeddings and how they can improve NLU performances---rather than an attempt 
to create state-of-the-art results.
Thus for example, in tagging tasks (e.g., NER, POS), we 
follow the protocols in \cite{collobert2011natural} using the 
concatenation of neighboring embeddings as input features rather
than treating embeddings as auxiliary features which are
fed into a CRF model along with other manually developed features
as in \newcite{pennington2014glove}.  Or for experiments on sentiment and other tasks 
where sentence level embeddings are required  we only
employ standard recurrent or recursive models for
sentence embedding rather than models with sophisticated state-of-the-art
methods (e.g., \newcite{tai2015improved,irsoy2014deep}).

Significance testing for comparing models is done via the bootstrap test \cite{efron1994introduction}.
Unless otherwise noted, significant testing is performed on one-word-one-vector embedding (50d) versus multi-sense embedding using Expectation  inference (50d) and one-vector embedding (100d) versus Expectation (100d).

\subsection{The Tasks}

\paragraph{Named Entity Recognition} 
We use the CoNLL-2003 English benchmark for training, and test on the CoNLL-2003 test data. 
We follow the protocols in \newcite{collobert2011natural},
using the concatenation of neighboring embeddings as input to a multi-layer neural model.
We employ a five-layer neural architecture, comprised of
an input layer, three convolutional layers with rectifier linear
activation function and a softmax output layer. Training is done
by gradient descent with minibatches where
each sentence is treated as one batch. Learning rate, window size,
number of hidden units of hidden layers, L2 regularizations and
number of iterations are tuned on the development set.
\begin{table}[!ht]
\small
\centering
\begin{tabular}{ccc}\hline
Standard (50)&Greedy (50)&Expectation( 50)\\
0.852&0.852 (+0)&0.854 (+0.02)\\\hline
Standard (100)&Global+G (100)&Global+E (100)\\
 0.867& 0.866 (-0.01) &0.871 (+0.04)\\\hline
 Standard (300)\\
  0.882\\\hline
\end{tabular}
\caption{Accuracy for Different Models on Name Entity Recognition.  Global+E stands for Global+Expectation inference and Global+G stands for Global+Greedy inference. p-value 0.223 for Standard(50) verse Expectation (50) and 0.310 for Standard(100) verse Expectation (100).}
\label{NER}
\end{table}

\paragraph{Part-of-Speech Tagging}  
We use  Sections 0–18 of the Wall Street Journal (WSJ) data for training, 
sections 19–21 for validation and sections 22–24 for testing. 
Similar to NER, we trained 5-layer neural models which take the
concatenation of neighboring embeddings as inputs.  We adopt a
similar training and parameter tuning strategy as for POS tagging.
\begin{table}[!ht]
\small
\centering
\begin{tabular}{ccc}\hline
Standard (50)&Greedy (50)&Expectation (50)\\
0.925&0.934 (+0.09)&0.938 (+0.13)\\\hline
Standard (100)&Global+G (100)&Global+E (100)\\
0.940&0.946 (+0.06) &0.952 (+0.12)\\\hline
 Standard (300)\\
  0.954\\\hline
\end{tabular}
\caption{Accuracy for Different Models on Part of Speech Tagging. P-value 0.033 for 50d and 0.031 for 100d.}
\label{POS}
\end{table}

\paragraph{Sentence-level Sentiment Classification (Pang)}
The sentiment dataset of \newcite{pang2002thumbs} consists of movie
reviews with a sentiment label for each sentence. 
We divide the original dataset into training(8101)/dev(500)/testing(2000).  
Word embeddings are initialized using the aforementioned types of embeddings
and kept fixed in the learning procedure. 
Sentence level embeddings are
achieved by using standard sequence recurrent neural models \cite{pearlmutter1989learning} (for details, please refer to Appendix section).  
The obtained embedding is  then fed into a sigmoid classifier.
Convolutional matrices at the word level 
are randomized from [-0.1, 0.1] and learned from sequence models. 
For training, we adopt 
AdaGrad with mini-batch. Parameters (i.e., 
$L2$ penalty, 
learning rate and  mini batch size) are tuned on the development set.
Due to space limitations, we omit details of recurrent models and training.
\begin{table}[!ht]
\small
\centering
\begin{tabular}{ccc}\hline
Standard (50)&Greedy (50)&Expectation (50)\\
0.750&0.752(+0.02)&0.750(+0.00)\\\hline
Standard (100)&Global+G (100)&Global+E (100)\\
0.768& 0.765(-0.03) &0.763(-0.05)\\\hline
 Standard (300)\\
  0.774\\\hline
\end{tabular}
\caption{Accuracy for Different Models on Sentiment Analysis (Pang et al.'s dataset). P-value 0.442 for 50d and 0.375 for 100d.}
\label{tagging}
\end{table}
\paragraph{Sentiment Analysis--Stanford Treebank}
The Stanford Sentiment Treebank \cite{socher2013recursive} contains gold-standard labels 
for each constituent in the parse tree (phrase level), thus allowing us to investigate
a sentiment task at a finer granularity than the
dataset in \newcite{pang2002thumbs} where labels are only found at the top of each sentence, 
\begin{comment}
The dataset includes gold standard labels for 215,154 phrases in parse trees
for 11,855 sentences, with an average sentence length of 19.1.  
\end{comment}
The sentences in the treebank were split into a
training(8544)/development(1101)/testing(2210) dataset.
\begin{comment}
The task has two settings, fine grained 
(5-class classification, very positive, positive, neutral, negative, very negative)
and coarse grained (binary positive vs negative classification).
\end{comment}

Following \newcite{socher2013recursive} we obtained embeddings for tree nodes 
by using a recursive neural  network model,
where the embedding for parent node is obtained in a bottom-up fashion based on its children. 
The embeddings for each parse tree constituent are output to a softmax layer;
see \newcite{socher2013recursive}.

We focus on the standard version of recursive neural models. 
Again we fixed word embeddings to each of the different embedding settings described
above\footnote{Note that this is different from the settings used in  \cite{socher2013recursive}  where
word vectors were treated as parameters to optimize.}. 
Similarly, we adopted AdaGrad with mini-batch. Parameters (i.e., 
$L2$ penalty, 
 learning rate and  mini batch size) are tuned on the development set.
The number of iterations is treated as a variable to tune and parameters are harvested based on the best performance on the development set. 
\begin{table}[!ht]
\small
\centering
\begin{tabular}{ccc}\hline
Standard (50)&Greedy (50)&Expectation (50)\\
0.818&0.815 (-0.03)&0.820 (+0.02)\\\hline
Standard (100)&Global+G (100)&Global+E (100)\\
0.838& 0.840 (+0.02) &0.838 (+0.00)\\\hline
 Standard (300)\\
  0.854\\\hline
\end{tabular}
\caption{Accuracy for Different Models on Sentiment Analysis (binary classification on Stanford Sentiment Treebank.). P-value 0.250 for 50d and 0.401 for 100d.}
\label{tagging}
\end{table}
\paragraph{Semantic Relationship Classification} SemEval-2010 Task 8 \cite{hendrickx2009semeval} is to find semantic relationships between
pairs of nominals, e.g., in ``My [apartment]$_{\text{e1}}$ has a pretty large [kitchen]$_{\text{e2}}$" 
classifying the relation between [apartment] and [kitchen] as {\it component-whole}. 
The dataset contains 9 ordered relationships, so the task is formalized as a 19-class classification problem,
with directed  relations treated as separate labels; see \newcite{hendrickx2009semeval} for details.

We follow the recursive implementations defined in \newcite{socher2012semantic}.    
The path in the parse tree between the two nominals is retrieved, and
the embedding is calculated based on recursive models and fed to a softmax classifier. 
For pure comparison purpose,
we only use embeddings as features and do not explore other combination of artificial features.
We adopt the same training strategy as for the sentiment task (e.g., Adagrad, minibatches, etc). 
\begin{table}[!ht]
\small
\centering
\begin{tabular}{ccc}\hline
Standard (50)&Greedy (50)&Expectation (50)\\
0.748&0.760 (+0.12)&0.762 (+0.14)\\\hline
Standard(100)&Global+G (100)&Global+E (100)\\
0.770& 0.782 (+0.12) &0.778 (+0.18)\\\hline
 Standard(300)\\
  0.798\\\hline
\end{tabular}
\caption{Accuracy for Different Models on Semantic Relationship Identification. P-value 0.017 for 50d and 0.020 for 100d.}
\label{tagging}
\end{table}
\paragraph{Sentence Semantic Relatedness} 
We use the Sentences Involving Compositional
Knowledge (SICK) dataset \cite{marelli2014semeval}
 consisting of 9927 sentence pairs, split into
training(4500)/development(500)/Testing(4927). 
Each sentence pair is associated with a gold-standard label ranging from 1 to 5,
indicating how semantically related are the two sentences, from
1 (the two sentences are unrelated) to 5 (the two are very related).

In our setting,
the similarity between two sentences is measured based on sentence-level embeddings. Let $s_1$ and $s_2$ denote two sentences and $e_{s_1}$ and $e_{s_2}$ denote corresponding embeddings. 
$e_{s_1}$ and $e_{s_2}$ are achieved through recurrent or recursive models (as illustrated in Appendix section).
Again, word embeddings are obtained by simple table look up in one-word-one-vector settings
 and inferred using the Greedy or Expectation strategy in multi-sense settings. 
We adopt two different recurrent models for acquiring sentence-level embeddings, 
a standard recurrent model and an LSTM model \cite{hochreiter1997long}. 

The similarity score is predicted using a regression model
built on the structure of a three layer convolutional model,
with concatenation of $e_{s1}$ and $e_{s2}$ as input, and a regression score from 1-5 as output.
We adopted the same training strategy as described earlier.  The trained
model is then used to predict the relatedness score between two new
sentences. Performance is measured using Pearson's $r$ between
the predicted score and gold-standard labels.
\begin{table}[!ht]
\small
\centering
\begin{tabular}{ccc}\hline
Standard( 50)&Greedy (50)&Expectation (50)\\
0.824&0.838(+0.14)&0.836(+0.12)\\\hline
Standard (100)&Global+G (100)&Global+E (100)\\
0.835& 0.840 (+0.05) &0.845 (+0.10)\\\hline
 Standard(300)\\
  0.850\\\hline
\end{tabular}
\caption{Pearson's $r$ for Different Models on Semantic Relatedness for Standard Models. P-value 0.028 for 50d and 0.042 for 100d.}
\label{tagging}
\end{table}
\begin{table}[!ht]
\small
\centering
\begin{tabular}{ccc}\hline
Standard(50)&Greedy(50)&Expectation(50)\\
0.843&0.848 (+0.05)&0.846 (+0.03)\\\hline
Standard(100)&Global+G (100)&Global+E (100)\\
0.850& 0.853 (+0.03) &0.854 (+0.04)\\\hline
 Standard(300)\\
  0.850\\\hline
\end{tabular}
\caption{Pearson's $r$ for Different Models on Semantic Relatedness for LSTM Models. P-value 0.145 for 50d and 0.170 for 100d.}
\label{tagging}
\end{table}

\subsection{Discussions}
Results for different tasks are represented in Tables 3-9. 

At first glance it seems  that
multi-sense embeddings do indeed offer superior performance, since
combining global vectors with sense-specific vectors introduces 
a consistent performance boost for every task, when compared with the standard (50d) setting.
But of course this is an unfair comparison; combining global vector with sense-specific vector  doubles
the dimensionality of vector to 100, making 
comparison with standard dimensionality (50d) unfair.
When comparing with standard (100), the conclusions become more nuanced.

For every task, the +Expectation method has performances that often seem to be
higher than the simple baseline (both for the 50d case or the 100d case).
However, only some of these differences are significant.

(1)
Using multi-sense embeddings is significantly helpful for
tasks like semantic relatedness (Tables 7-8). 
This is sensible since
sentence meaning here is sensitive to the semantics of one particular
word, which could vary with word sense and which would directly be reflected on the  relatedness score. 

(2) By contrast, for sentiment analysis (Tables 5-6), much of the task depends on correctly
identifying a few sentiment words like ``good" or ``bad",
whose senses tend to have similar sentiment values, and hence for which
multi-sense embeddings offer little help.
Multi-sense embeddings might promise to help sentiment analysis for some cases, like
disambiguating the word ``sound" in ``safe and sound'' versus ``movie sound".
But we suspect that such cases are not  common, explaining the non-significance of the improvement.
Furthermore, the advantages of neural models in  sentiment analysis tasks 
presumably lie in their capability to capture local composition like negation,
and it's not clear how helpful multi-sense embeddings are for that aspect.

(3) Similarly, multi-sense embeddings help for POS tagging, but not
for NER tagging (Table 3-4).   Word senses have long been  known to be related to POS tags.
But the largest proportion of NER tags consists of the negative not-a-NER (``O") tag, 
each of which is likely correctly labelable regardless 
of whether senses are disambiguated or not (since presumably if a word is not a named entity, 
most of its senses are not named entities either).

(4) As we apply more sophisticated models like LSTM to semantic relatedness tasks (in Table 9), 
the advantages caused by multi-sense embeddings disappears.

%Nonetheless, despite these trends toward a difference in absolute range, it turns out that
%for every task the difference between using
%multi-sense embeddings and using the standard skip-gram baselines of equivalent dimensionality
%is {\bf not} statistically significant by the bootstrap test.

(5) Doubling the number of dimensions is sufficient to increase
performance as much as using the complex multi-sense algorithm.
(Of course increasing vector dimensionality (to 300) boosts performance even more,
although at the significant cost of exponentially increasing time complexity.)
We do larger one-word-one-vector embeddings do so well?
We suggest some hypotheses:
\begin{itemize}
\item though information about distinct senses is encoded in
one-word-one-vector embeddings in a mixed and
less structured way, we suspect that the compositional nature of neural models is able to 
separate the informational chaff from the wheat and choose
what information to take up, bridging the gap between
single vector and multi-sense paradigms. 
For models like LSTMs which are better at doing such a job by
using gates to control information flow, the difference
between two paradigms should thus be further narrowed, as indeed we found.
\item
The pipeline model proposed in the work requires sense-label inference (i.e., step 2). We proposed two strategies: \textsc{greedy} and \textsc{expectation}, and found that
\textsc{greedy} models perform worse than \textsc{expectation}, as we might expect\footnote{ 
\textsc{greedy} models work in a more aggressive way and likely make mistakes due to the non-global-optimum nature and 
limited context information}.  
But even \textsc{expectation} can be viewed as another form of one-word-one-vector models,
just one where different senses are entangled but weighted to emphasize the important ones.
Again, this suggests another cause for the strong relative performance of larger-dimensioned one-word-one-vector models.
\end{itemize}

%(p-value 0.207 for 50d Expectation VS Standard). 
%
%Similar explanations can be applied to POS and NER tasks.
%Disambiguating sense embeddings potentially  induces clear evidence for POS tagging but it is less beneficial for NER:
%
%Additionally, some specific types of words like propositions play important roles in NER and  disambiguating word senses will not necessarily be of help with that respect. 
%

\section{Conclusion}
In this paper, we expand ongoing research into multi-sense embeddings
by first proposing a new version based on Chinese restaurant processes
that achieves state of the art performance on simple word similarity matching tasks.
We then introduce a pipeline system for incorporating multi-sense embeddings into NLP applications,
and examine multiple NLP tasks to see whether and when multi-sense embeddings can introduce 
performance boosts.  Our results suggest that simply increasing the dimensionality
of baseline skip-gram embeddings is sometimes sufficient to achieve the same performance wins that come from  using
multi-sense embeddings.
That is, the most straightforward way to yield better
performance on these tasks is just to increase embedding dimensionality.

Our results come with some caveats.  
In particular, our conclusions are based on the pipelined system
that we introduce, and other multi-sense embedding systems (e.g., a more
advanced sense learning model or a better sense label model or a
completely different pipeline system) may find stronger effects of multi-sense models.
Nonetheless
we do consistently find improvements for multi-sense embeddings in some tasks
(part-of-speech tagging and  semantic relation identification), suggesting the benefits of our
multi-sense models and those of others.
Perhaps the most important implication of our results may be the evidence they
provide for the importance of going beyond simple human-matching tasks,
and testing embedding models by using them as components in real NLP applications.

%We explore  the behavior and results of multi-sense embeddings in different scenarios.  
%There are also a couple of points we wish to emphasize here:

\section{Appendix}
In sentiment classification and sentence semantic relatedness tasks,  
classification models require embeddings that represent the input at a  sentence or phrase level.
We adopt recurrent networks (standard ones or LSTMs) and recursive networks  
in order to map a sequence of tokens with various length to a vector representation.

\paragraph{Recurrent Networks} 
A recurrent network successively takes word $w_t$ at step $t$,
combines its vector representation $e_{t}$ with the previously built
hidden vector $h_{t-1}$  from time $t-1$, calculates the resulting
current embedding $h_t$, and passes it to the next step.  The embedding
$h_t$ for the current time $t$ is thus:
\begin{equation}
h_{t}=\text{tanh}(W\cdot h_{t-1}+V\cdot e_{t})
\end{equation}
where $W$ and $V$ denote compositional matrices. 
If $N_s$ denote the length of the sequence, $h_{N_s}$ represents
the whole sequence $S$. 
\paragraph{Recursive Networks}
Standard recursive models work in a similar way by working on neighboring words
by parse tree order rather than sequence order.
They compute the
representation for each parent node based on its immediate
children recursively in a bottom-up fashion until reaching the root
of the tree. For a given node $\eta$ in the tree and
its left child
$\eta_{\text{left}}$ (with  representation $e_{{\text{left}}}$) and right 
child $\eta_{\text{right}}$ (with  representation $e_{{\text{right}}}$),
the standard recursive network calculates $e_{\eta}$:
\begin{equation}
e_{\eta}=\text{tanh}(W\cdot e_{\eta_{\text{left}}}+V\cdot e_{\eta_{\text{right}}})
\end{equation}

\paragraph{Long Short Term Memory (LSTM) }
LSTM models \cite{hochreiter1997long} are defined as follows: 
given a sequence of inputs  $X=\{x_1,x_2,...,x_{n_X}\}$, an LSTM associates each timestep with an input, memory and output gate, 
respectively denoted as $i_t$, $f_t$ and $o_t$.
We notationally disambiguate $e$ and $h$, where $e_t$ denote the vector for an individual text unit (e.g., word or sentence) at time step t 
while $h_t$ denotes the vector computed by the LSTM model at time t by combining $e_t$ and $h_{t-1}$. 
$\sigma$ denotes the sigmoid function. 
$W\in \mathbb{R}^{4K\times 2K}$.
The vector representation $h_t$ for each time-step $t$ is given by:

\begin{equation}
\Bigg[
\begin{array}{lr}
i_t\\
f_t\\
o_t\\
l_t\\
\end{array}
\Bigg]=
\Bigg[
\begin{array}{c}
\sigma\\
\sigma\\
\sigma\\
\text{tanh}\\
\end{array}
\Bigg]
W\cdot
\Bigg[
\begin{array}{c}
h_{t-1}\\
e_{t}\\
\end{array}
\Bigg]
\end{equation}
\begin{equation}
c_t=f_t\cdot c_{t-1}+i_t\cdot l_t\\
\end{equation}
\begin{equation}
h_{t}^s=o_t\cdot c_t
\end{equation}
\section{Acknowledgments}
We would  like to thank
Sam Bowman, Ignacio Cases, Kevin Gu,  Gabor Angeli, Sida Wang, Percy Liang 
and other members of the Stanford NLP group, 
as well as anonymous reviewers for their helpful advice on various aspects of this work. 
We gratefully acknowledge the support of the NSF via award IIS-1514268, the Defense Advanced Research Projects Agency (DARPA) Deep Exploration and Filtering of Text (DEFT) Program under Air Force Research Laboratory (AFRL) contract no. FA8750-13-2-0040.
Any opinions, findings, and conclusions or recommendations expressed in this material are those of the authors and do not necessarily reflect the views of NSF, DARPA, AFRL, or the US government.

\bibliographystyle{acl}
\bibliography{acl2013}  
\end{document}